# Applications of Binary Similarity and Distance Measures

Manoj Muniswamaiah, Tilak Agerwala and Charles C. Tappert
Seidenberg School of CSIS, Pace University, White Plains, New York
{mm42526w, tagerwala, ctappert}@pace.edu

**Abstract** In the recent past, binary similarity measures have been applied in solving biometric identification problems, including fingerprint, handwritten character detection, and in iris image recognition. The application of the relevant measurements has also resulted in more accurate data analysis. This paper surveys the applicability of binary similarity and distance measures in various fields.

**Keywords**: binary similarity, clustering, binary measures, classification.

## I. Introduction

Similarity and distance (dissimilar) measures are essential in solving problems in pattern recognition. These measures include clustering data points based on how similar, and distant (dissimilar) they are [1][2]. Similarity and distance measures are conducted using methods that use different mathematical approaches in measuring how near or far two objects may be in their respective locations. Mathematical approaches that measure similarity include cosine similarity that measures similarity irrespective of the object's sizes. Cosine similarity works by calculating the cosine of the angle between the two objects projected in multi-dimensional space; the Manhattan distance that measures similarity based on the absolute difference between the Cartesian coordinates of the two objects; and the Euclidean distance that measures the length of the segment that joins two points of the objects.

The binary similarity distant measure is a method used in calculating similarity and distance in binary variables. Hence, the objects whose patterns need to be discovered must possess discreet values, providing for only two atomic results (yes/no; [1,0]). This paper highlights some of these mathematical formulas used in achieving the binary similarity and distance measure, as well as respective areas where these formulas are applied.

## II. Related Work

The suitability of using the binary similarity and distance measure was tested in a smart home setup, where the researchers tested the method's effectiveness in pattern recognition for applications including image retrieval and character recognition. Data was collected from two sources: the door entry point and the lounge area. The binary similarity distance measure was then used to measure the degree of resemblance and difference between the movements recorded in these two areas. Only two sequence values were considered in this test: the positive match and mismatch [3].

Jaccard similarity measures have always been used for clustering of biological species, and Forbesi gave a proposal of clustering coefficient for the measurement of related species. Subsequently, binary similarity measures have been applied for ethnology, taxonomy, and geology. Still, only a few studies have been conducted on binary distance, and similarity measurements. Classification and statistical equations on binary similarity and distance measures have been applied to detect anomalies in various fields. The goal of anomaly detection and network intrusion is to establish whether the specific observations (network activities) are anomalous or not. Anomaly detection needs the labeling of observations for it to be consistent and effective. It involves various features which are used in the observation, featuring and identification of the datasets. [1][2][3]

| Set   | Character | 1                          | 0                              |
|-------|-----------|----------------------------|--------------------------------|
| Set 1 | 1         | a = 1/1 (in both Sets)     | b = 1/0 (only in Set 1)        |
| Set 2 | 0         | c = 0/1 (only in Set 2)    | d = 0/0 (in none of the Sets)  |

Above table represents variables used to calculate binary similarity and distance measures

# III. Application of Binary Similarity and Distance Measures

| Binary Similarity & Distance Measure | Equation | Application |
|---|---|---|
| 1. The Manhattan distance | MD(x,y)=$\sum_{i=1}^{n}\|x_i - y_i\|$ [1][2] | Regression analysis [1] |
| 2. Chebyshev distance | CD(x,y)=$\max_i\|x_i - y_i\|$ [2] | Warehouse logistics [2] |
| 3. Euclidean (ED) | ED(x,y)=$\sqrt{\sum_{i=1}^{n}\|x_i - y_i\|^2}$ [2][3] | The Euclidean Distance (ED) is used to identify genes [3] |
| 4. L1 Distance measures | LD(x,y)= $\sum_{i=1}^{n} ln(1 + \|x_i - y_i\|)$, [2][4] | It is used in image retrieval [4] |
| 5. Canberra distance (CanD) | CanD(x,y)=$\sum_{i=1}^{n} \frac{\|x_i - y_i\|}{\|x_i\|+\|y_i\|}$ [2][5] | Intrusion detection in computer systems and comparison of ranked lists [5] |
| 6. The Sorensen distance | SD(x,y)=$\frac{\sum_{i=1}^{n}\|x_i-y_i\|}{\sum_{i=1}^{n}(x_i+y_i)}$ [2][6] | Computer lexicography where it measures the lexical relationship score between two particular words [6] |
| 7. Soergel distance | SoD(x,y)=$\frac{\sum_{i=1}^{n}\|x_i-y_i\|}{\sum_{i=1}^{n} max(x_i,y_i)}$ [2][7] | Applied in calculation of evolutionary distances [7] |
| 8. Kulczynski distance (KD) | KD(x,y)=$\frac{\sum_{i=1}^{n}\|x_i-y_i\|}{\sum_{i=1}^{n} min(x_i,y_i)}$ [2][8] | Applied in ecology to determine phylogenetic and geographic variability in bacterial distribution on plant leaves [8] |
| 9. Mean Character Distance (MCD) | MCD(x,y)= $\frac{\sum_{i=1}^{n}\|x_i-y_i\|}{n}$ [2][9] | Used to determine the linkage equilibrium [9] |
| 10. Non-Intersection Distance (NID) | NID(x,y)=$\frac{1}{2}\sum_{I=1}^{N}\|X_I - Y_I\|$. [2][10] | Developing protocol for identifying crash risk areas on GIS for pedestrians [10] |
| 11. Jaccard distance (JacD) | JacD=$\frac{\sum_{i=1}^{n}(x_i-y_i)^2}{\sum_{i=1}^{n} x_i^2+\sum_{i=1}^{n} y_i^2-\sum_{i=1}^{n} x_i y_i}$ [2][11] | Applied in measuring distance of tumor evolutionary trees [11] |
| 12. Cosine distance (CosD) | CosD(x,y)=1-$\frac{\sum_{i=1}^{n} x_i y_i}{\sqrt{\sum_{i=1}^{n} x_i^2}\sqrt{\sum_{i=1}^{n} y_1^2}}$ [2][12] | Applied in test analysis to measure similarities in documents [12] |
| 13. Dice distance (DicD) | DicD(x,y)=1-$\frac{2\sum_{i=1}^{n} x_i y_i}{\sum_{i=1}^{n} x_i^2+\sum_{i=1}^{n} y_i^2}$ [2][13] | Applied in gene measurement and analysis [13] |
| 14. Chord distance (ChoD) | $ChoD(x,y) = \sqrt{2 - 2\frac{\sum_{i=1}^{n} x_i y_i}{\sum_{i=1}^{n} x_i^2 \sum_{i=1}^{n} x_i^2 \sum_{i=1}^{n} y_i^2}}$ [2] [14] | Quantification and representation of diverse polycrystalline microstructures [14] |
| 15. Bhattacharyya distance (BD) | BD(x,y)=-$\ln\sum_{i=1}^{n} \sqrt{x_i y_i}$ [2][15] | Applied in studying extraction and selection of feature, phone clustering, image processing and speaker recognition [15] |
| 16. Squared chord distance (SCD) | SCD(x,y)=$\sum_{i=1}^{n}(\sqrt{x_i} - y_i)$ [2][14] | Applied in quantification and representation of diverse polycrystalline microstructures [14] |
| 17. Matusita distance (MatD) | MatD(x,y)=$\sqrt{\sum_{i=1}^{n}(\sqrt{x_1} - \sqrt{y_i})^2}$ [2][16] | Applied in optimal band selection and evaluation of classification results [16] |
| 18. Hellinger distance (HeD) | HeD(x,y)=$\sqrt{2\sum_{i=1}^{n}(\sqrt{x_1} - \sqrt{y_i})^2}$ [2][17] | Applied in classification of passive underwater acoustic signals [17] |
| 19. Squared Euclidean distance (SED) | $SED(x,y) = \sum_{i=1}^{n}(x_i - y_i)^2$ [2][18] | Applied in least square methods to fit statistical estimates to data [18] |
| 20. Clark distance (ClaD) | $ClaD(x,y) = \sqrt{\sum_{i=1}^{n}\left(\frac{|x_i - y_i|}{x_i + y_i}\right)^2}$ [2][19] | Applied in solving vehicle routing problem [19] |
| 21. Neyman χ2 distance (NCSD) | $NCSD(x,y) = \sum_{i=1}^{n}\frac{(x_i - y_i)^2}{x_i}$ [2][20] | Applied in medical statistics to test homogeneity [20] |
| 22. Pearson χ2 distance (PCSD) | $PCSD(x,y) = \sum_{i=1}^{n}\frac{(x_i - y_i)^2}{y_i}$ [2][20] | Applied in genetic studies to test homogeneity, and goodness of fit [20] |
| 23. Squared χ2 distance (SquD) | $SquD(x,y) = \sum_{i=1}^{n}\frac{(xi-yi)^2}{xi+yi}$ [2][20] | Used in measuring blind weight disparity [20] |
| 24. Probabilistic Symmetric χ2 distance (PSCSD) | $PSCSD(x,y) = 2\sum_{i=1}^{n}\frac{(xi-yi)^2}{xi+yi}$ [2][21] | Applied in testing dissimilarities in ecology and image processing problems [21] |
| 25. Mean Censored Euclidean Distance (MCED) | $MCED(x,y) = \sqrt{\frac{\sum_{i=1}^{n}(xi-yi)^2}{\sum_{i=1}^{n}1_{x_i^2+y_i^2 \neq 0}}}$ [2][22] | Used in performing dissimilarity measure for developing prognostic system for cancer patients [22] |
| 26. Squared Chi-Squared (SCSD) | $SCSD(x,y) = \sum_{i=1}^{n}\frac{(xi-yi)^2}{|xi+yi|}$ [2][20] | Used in analyzing group differences when the dependent variable is measured at a nominal level [20] |
| 27. Average distance (AD) | $AD(x,y) = \sqrt{\frac{1}{n}\sum_{i=1}^{n}(xi-yi)^2}$ [2][23] | Used in clustering a larger number of data set using the SOTA (self-organizing tree algorithm) approach [23] |
| 28. Additive Symmetric χ2 (ASCSD) | $ASCSD(x,y) = 2\sum_{i=1}^{n}\frac{(xi-yi)^2(xi+yi)}{xiyi}$ [2][20] | Applied is solving clustering problems [20] |
| 29. Divergence distance (DivD) | $DivD(x,y) = 2\sum_{i=1}^{n}\frac{(xi-yi)^2}{(xi+yi)^2}$ [2][24] | Used in hypothesis testing and econometric estimation [24] |
| 30. Kullback-Leibler distance (KLD) | $KLD(x,y) = \sum_{i=1}^{n}xi\,ln\frac{xi}{yi}$ [2][25] | Applied in characterizing the relative (Shannon) entropy in information systems, randomness in continuous time-series, and information gain when comparing statistical models of inference [25] |
| 31. Jeffreys Distance (JefD) | $JefD(x,y) = \sum_{i=1}^{n}(xi - yi)\,ln\frac{xi}{yi}$ [2][24] | Applied in performing tests and likelihood ratios [24] |
| 32. K divergence Distance (KDD) | $KDD(x,y) = \sum_{i=1}^{n}xi\,ln\frac{2xi}{xi+yi}$ [2][20] | Used in solving partitioning-based clustering problems [20] |
| 33. Topsoe Distance (Top) | $TopD(x,y) = \sum(xi\,ln(2xi/(xi+yi)) + yi\,ln(2yi/(xi+yi)))$ [26][2] | Applied in quantifying the distance measures and similarities of vectors by transposing the distance between two probability distributions [26] |
| 34. Jensen-Shannon Distance (JSD) | $JSD(x,y) = \sum(xi\,ln(2xi/(xi+yi)) + yi\,ln(2yi/(xi+yi)))/2$ [27][2] | JSD is applied in carrying out comparisons of protein surface, genome, bioinformatics, and in quantitative studies of history [27]. |
| 35. Vicis-Wave Hedges distance (VWHD) | $VWHD(x,y) = \sum_{i=1}^{n}\frac{|xi-yi|}{min(xi,yi)}$ [2][28] | VWHD has been applied in retrieving compressed image, retrieving content-based video, fingerprint recognition, time series classification, and image fidelity [28] |

| | | |
|---|---|---|
| 36. Vicis symmetric distance (VSD) | $VSDF1(x,y) = \sum_{i=1}^{n} \frac{(xi-yi)^2}{min\,(xi,yi)^2}$, [2][20] $VSDF2(x,y) = \sum_{i=1}^{n} \frac{(xi-yi)^2}{min\,(xi,yi)}$, [2][20] $VSDF2(x,y) = \sum_{i=1}^{n} \frac{(xi-yi)^2}{max\,(xi,yi)}$ [2][20] | Applied in torque, position, and speed control in intelligence control systems [20] |
| 37. Max symmetric χ2 distance (MSCD) | MSCD (x, y) = max (∑(xi − yi)2/xi, ∑(xi − yi)2/yi) [29][2] | Applied in regularized robust estimation of binary regression models [29] |
| 38. Min symmetric χ2 distance (MiSCSD) | MiSCSD (x,y) = min (∑(xi − yi)2/xi, ∑(xi − yi)2/yi) [20][2] | Applied in distance measures for the classification of numerical features [20] |
| 39. Average (L1, L∞) distance (AvgD) | $AvgD(x,y) \frac{\sum_{i=1}^{n}|xi-yi|+maxi|xi-yi|}{2}$ [20][2] | Applied in developing algorithm for solving single facility location problem [20] |
| 40. Kumar- Johnson Distance (KJD) | KJD(x,y)=∑(x2i-y2i )2/2(xiyi)3/2[30][2] | KJD is applied in developing a model for wave attenuation for mangrove farming [30] |
| 41. Taneja Distance (TanD) | TJD(x,y)=∑(xi+yi)/2 ln((xi+yi)/(2√xiyi)) [20][2] | Used in grouping similarity measures by caveats to implementation for probability density function [20][31][32] |
| 42. Hamming Distance (HamD) | $HamD(x.y) = \sum_{i=1}^{n} 1_{xi \neq yi}$ [2][33] | Applied in genetic studies to develop a model that recognizes DNA [33] |
| 43. Hausdorff Distance (HauD) | $HauD(x,y) = max(h(x,y),h(y,x))$[34][2] | Hausdorff distance is applied in carrying out comparison of meshes in simplification algorithms [34] |
| 44. χ2 statistic Distance (CSSD) | $CSSD(x,y) = \sum_{i=1}^{n} \frac{xi-mi}{mi}$ [35][2] | Applied in developing subspecialist management models to improve the quality control [35] |
| 45. Whittaker's index of association Distance (WIAD) | $WIAD(x,y) = \frac{1}{2}\sum_{i=1}^{n}\left|\frac{xi}{\sum_{i=1}^{n}xi} - \frac{yi}{\sum_{i=1}^{n}yi}\right|$ [2][20] | WIAD is applied in measurement of biodiversity of species, especially aquatic species [20] |
| 46. Meehl Distance (MeeD) | $MeeD(x,y) = \sum_{i=1}^{n=1}(xi - yi - xi + 1 + yi + 1)$ [2][36] | Is applied in determining accuracy and consistency in statistical prediction [36] |
| 47. Motyka Distance (Mot) | $MotD(x,y) = \frac{\sum_{i=1}^{n} max\,(xi,yi)}{\sum_{i=1}^{n}(xi+yi)}$ [2] [37] | Is applied in author profiling to determine demographics about a particular document and author [37] |
| 48. Hassanat Distance (HasD) | $HasD(xi,y) = \sum_{i=1}^{n} D(xi,yi)$ [2][38] | Is applied in machine learning to enhance the performance of the nearest neighbors' classifiers [38][39] |
| 49. Minkowski Distance | (b + c)^{1/1} [1] [43] | The Minkowski distance has been applied in modelling genetic algorithm for diagnostic of diabetes [43] |
| 50. Mountford Similarity | $\frac{a}{0.5(ab+ac)+bc}$ [1][44] | Applied in the comparison of the structural units of biocenosis consisting of the considered flora and the parasitic fauna [44] |
| 51. Tarwid Similarity | $\frac{na-(a+b)(a+c)}{na+(a+b)(a+c)}$ [1][20] | Applied in classification of genes in medical practice [20] |
| 52. Kulczynski-II Similarity | $\frac{\frac{a}{2}(2a+b+c)}{(a+b)(a+c)}$ [1][45] | Applied in analysis of biotic elements where species with small distribution areas are grouped [45] |
| 53. Johnson Similarity | $\frac{a}{a+b} + \frac{a}{a+c}$ [1][20] | Applied in determining the minimum distance between convex sets [20] |
| 54. Simpson Similarity | $\frac{a}{min\,(a+b,a+c)}$ [1][46] | Is applied in calculating areas in surveying [46] |
| 55. Peirce Similarity | $\frac{ab+bc}{ab+2bc+cd}$ [1][20] | Applied in learning and detecting content of languages [20] |

| | | | |
|---|---|---|---|
| 56. Tarantula Similarity | $\dfrac{\dfrac{a}{(a+b)}}{\dfrac{c}{(c+d)}} = \dfrac{a(c+d)}{c(a+b)}$ [1][47] | | Is applied in clustering high dimensional data set [47] |
| 57. Cole Similarity | $\dfrac{\sqrt{2}(ad-bc)}{\sqrt{(ad-bc)^2 - (a+b)(a+c)(b+d)(}}$ [1][20] | | Is widely applied in image comparison, image segmentation, and document clustering [20] |
| 58. 3W−Jaccard Similarity | $\dfrac{3a}{3a+b+c}$ [1][49] | | Has been used in diagnosis of dental and oral diseases using method of case-based reasoning. It measures higher common features [49] |
| 59. Nei & Li Similarity | $\dfrac{2a}{(a+b)(a+c)}$ [1][50] | | Is used in measuring spatial locations of soil [50] |
| 60. Sokal & Sneath−I Similarity | $\dfrac{2a}{a+2b+2c}$ [1][50] | | Is used in 3D mesh segmentation in computer vision [50]. |
| 61. Sokal & Michener Similarity | $\dfrac{a+d}{a+b+c+d}$ [1][51] | | Used in handwriting identification [51] |
| 62. Roger & Tanimoto Similarity | $\dfrac{a+d}{a+2(b+c)+d}$ [1][52] | | Mostly used within the medicinal field including similarity-based fingerprint calculations [52] |
| 63. FaITH Similarity | $\dfrac{a+0.5d}{a+b+c+d}$ [1][53] | | Is been applied in ontology concepts [53] |
| 64. Gower & Legendre Similarity | $\dfrac{a+d}{a+0.5(b+c)+d}$ [1][54] | | Has been extended to ordinal data types for purposes of multi-dimensional scaling and cluster analysis [54] |
| 65. Russel & Rao Similarity | $\dfrac{a}{a+b+c+d}$ [1][57] | | Has been applied in cluster analysis of dichotomous data. [57] |
| 66. Mean−Manhattan distance | $\dfrac{b+c}{a+b+c+d}$ [1][58] | | Is commonly used to measure the distance between images [58] |
| 67. Cityblock distance | b + c [1][59] | | Is used in distinguishing the distance between two points in a city. Also known as taxicab geometry. [59] |
| 68. Vari distance | $\dfrac{b+c}{4(a+b+c+d)}$ [1][60] | | Is used in probability measures including filtered measure [60] |
| 69. Sizedifference distance | $\dfrac{(b+c)2}{(a+b+c+d)2}$ [1][61] | | Is used in measurements [61] |
| 70. Shapedifference distance | $\dfrac{n(b+c)-(b-c)2}{(a+b+c+d)2}$ [62][1] | | Is used in scenarios where shape of an object is used to define characteristics [62] |
| 71. Patterndifference distance | $\dfrac{4bc}{(a+b+c+d)2}$ [63][1] | | Is used in combinatorial pattern matching, and spatial pattern matching. Combinatorial pattern matching is used in applications including DNA pattern matching, string matching, tree pattern |

| | | matching, as well as edit distance computation. Spatial pattern matching is used in finding a match between two images of given intensities or geometric transformation [63]. |
|---|---|---|
| 72. Lance & Williams distance | $$\frac{(b+c)}{2a+b+c}$$ [64][1] | Has been applied in kernel functions, as well as in designing threshold strategies for proximity values [64] |
| 73. Bray & Curtis distance | $$\frac{(b+c)}{2a+b+c}$$ [1][65] | Is used in identifying biological heterogeneities among the species [65]. |
| 74. Gilbert & Wells Similarity | $$\log a - \log n - \log\frac{a+b}{n} - \log\frac{a+c}{n}$$ [66][1] | Within ecology the algorithm is applied in classifying vegetation A rule set is made that helps in constructing plant vegetation types, based on specified combinations from a generic set of morphological and adaptive plant functioning attributes [66] |
| 75. Ochiai−I Similarity | $$\frac{a}{\sqrt{(a+b)(a+c)}}$$ [67][1] | Used in statistical analysis. [67] |
| 76. Forbesi Similarity | $\frac{na}{(a+b)(a+c)}$ [1][68] | Has been used in checking fingerprints. [68]. |
| 77. Fossum Similarity | $$\frac{n(a-o.5)2}{(a+b)(a+c)}$$ [1][69] | Has been used in 3D similarity searching that includes chemical structures. [69] |
| 78. Sorgenfrei Similarity | $$\frac{a2}{(a+b)(a+c)}$$ [1][70] | Is used in statistically analyzing alarm signals. [70] |

## IV. Conclusion

In a variety of domains, binary similarity and distance measurements have been applied. Each one has its own set of synthetic qualities that define it. Some use a simple count difference, while others use a complex correlation. The study anticipates that understanding the link between each set of measures will aid researchers in selecting more precise measures for binary data analysis in a variety of disciplines.